# Method for Hybrid Precision Convolutional Neural Network Representation

M. Al-Hami,  M. Pietron, R. Kumar, R.A. Casas, S.L. Hijazi, C. Rowen

## Problem and Solution

### Problem

This invention addresses fixed-point representations of convolutional neural networks (CNN) in integrated circuits.  When quantizing a CNN for a practical implementation there is a trade-off between the precision used for operations between coefficients and data and the accuracy of the system.  A homogenous representation may not be sufficient to achieve the best level of performance at a reasonable cost in implementation complexity or power consumption.  Parsimonious ways of representing data and coefficients are needed to improve power efficiency and throughput while maintaining accuracy of a CNN.

### Solution

Our invention provides means to represent coefficients and data of a CNN in a hybrid fashion by assigning fixed-point formatting to different partitions of convolutional layers and feature maps.  For instance, different precisions (i.e. numbers of bits) and different formats (i.e. numbers of integer and fractional bits) can be assigned to each two-dimensional (2D) kernel or three-dimensional (3D) kernel of a convolutional layer.  Similarly, different precisions and formats can be assigned to different partitions of the input and/or output feature maps.  While defining precisions and formats in a hybrid fashion adds overhead for representation both in terms of storage space and facilities to decode and use the information, it can bring about a significant improvement in CNN accuracy when compared to a homogeneous quantization (i.e. 4D).  Additionally, precisions for intermediate accumulated values can also be specified.

Thus, a CNN layer with 2D quantization could have representation as detailed below and as illustrated in Figure 1.

```
layer {
    name: "conv1"
    number of input channels: 2
    number of output channels: 3
    kernel size: 5x5 spatial kernels
    quantization structure: 2D
    coeff_precision: 8b
    coeff_format[1:5, 1:5, 0, 0]: <1:-5>    % coefficient format[x, y, input channel, output channel]
    coeff_format[1:5, 1:5, 0, 1]: <2:-4>    % <msb:lsb> (sign bit not shown)
    coeff_format[1:5, 1:5, 0, 2]: <2:-4>    % Example: <2:-4> ➔ <sgn  $b_2$ $b_1$ $b_0$ . $b_{-1}$ $b_{-2}$ $b_{-3}$ $b_{-4}$> ➔ 8b
```

```
        coeff_format[1:5, 1:5, 1, 0]: <5:-1>
        coeff_format[1:5, 1:5, 1, 1]: <6:0>
        coeff_format[1:5, 1:5, 1, 2]: <-1:-7>

        input_data_precision: 8b
        input_data_format[0]: <3:-3>    % input data format[input channel]
        input_data_format[1]: <4:-2>

        accumulator_precision: 16b
        accumulator_format:  <11:-3>

        output_data_precision: 8b
        output_data_format[0]: <10:4>    % output data format[output channel]
        output_data_format[1]: <11:5>
        output_data_format[2]: <9:3>
        …..
        …..
}
```

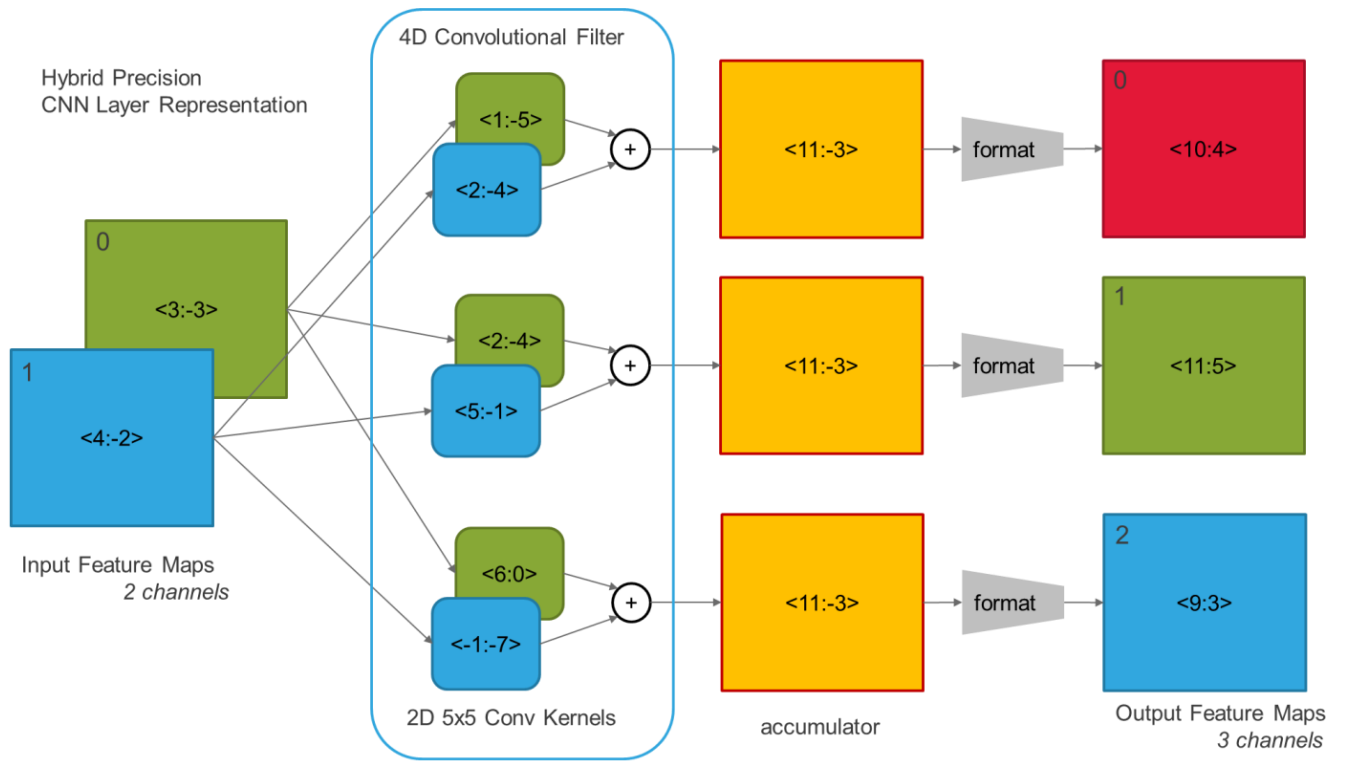

**Figure 1.  Hybrid precision representation in diagram form.**

# Prior Art

## Caffe

Caffe is a popular tool for testing and training neural networks [2]. It uses a *.prototxt file format to describe networks that does not include any information about fixed-point representations.

## Ristretto

Ristretto is a software tool used to define CNN fixed-point representations [2]. In the example below from the Ristretto website (http://lepsucd.com/?page_id=637) a single format (highlighted section) is used represent the coefficients and the output data for the layer; this would be considered a 4D quantization. Our work provides a more general representation by assigning a different format to each 2D or 3D convolutional kernel within the layer.

1. layer {
   1. name: "conv1"
   2. type: "ConvolutionRistretto"
   3. bottom: "data"
   4. top: "conv1"
   5. convolution_param {
      1. num_output: 96
      2. kernel_size: 7
      3. stride: 2
      4. weight_filler {
         5. type: "xavier"
         6. }
      }
   6. }
   7. quantization_param {
   8. precision: MINIFLOAT
   9. mant_bits: 10
   10. exp_bits: 5
   11. }
2. }

# Results

The table below shows the Top-1 validation error for AlexNet for different 8b fixed-point quantizations. It demonstrates that applying a naïve heterogeneous quantization to an entire layer (i.e. 4D) can lead to poor performance versus applying different representations to each 3D or 2D filter.

| 8b Quantization | AlexNet Top-1 Error [%] |
|---|---|
| 2D | 42.7 |
| 3D | 42.4 |
| 4D | 61.5 |

# Central Claim

1. A computer implemented method for representing a convolutional layer comprising:

    defining a first format for a first convolutional kernel that operates on a first feature map; and

    defining a second format for a second convolutional kernel that operates on a second feature map.